\documentclass[runningheads]{llncs}
\usepackage{graphicx}
\usepackage{hyperref} 
\usepackage{amssymb}
\usepackage{array}
\usepackage{graphicx}
\usepackage{subcaption}
\usepackage{booktabs}
\usepackage{url}
\usepackage{algorithm}
\usepackage{algpseudocode}
\usepackage{mathtools}
\usepackage{amsmath}
\usepackage{blindtext}
\usepackage{enumitem}
\usepackage{xcolor}
\usepackage{bm}

\newcommand{\xl}[1]{{\color{black}#1}}
\begin{document}
\title{Graph Neural Network for Interpreting Task-fMRI Biomarkers}

\author{Xiaoxiao Li$^{1}$, Nicha C. Dvornek$^{5}$, Yuan Zhou$^{5}$, Juntang Zhuang$^{1}$, \\Pamela Ventola$^{4}$ and James S. Duncan$^{1,2,3,5}$}

% 1{Li, Xiaoxiao}
% 2{Dvornek, Nicha}
% 3{Zhou, Yuan}
% 4{Zhuang, Juntang}
% 5{Ventola, Pamela}
% 6{Duncan, James}

\authorrunning{X. Li et al.}


\institute{$^{1}$ Biomedical Engineering, Yale University, New Haven, CT, USA \\
	$^{2}$ Electrical Engineering, Yale University, New Haven, CT, USA\\
	$^{3}$Statistics \& Data Science, Yale University, New Haven, CT, USA\\
	$^{4}$ Child Study Center, Yale School of Medicine, New Haven, CT, USA\\
	$^{5}$Radiology \& Biomedical Imaging, Yale School of Medicine, New Haven, CT, USA
}
\maketitle              % typeset the header of the contribution
%
% \footnotetext[1]{This work was supported by NIH Grant R01NS035193.}

\begin{abstract}
Finding the biomarkers associated with ASD is helpful for understanding the underlying roots of the disorder and can lead to earlier diagnosis and more targeted treatment. A promising approach to identify biomarkers is using \xl{Graph Neural Networks (GNNs)}, which can be used to analyze graph structured data, i.e. brain networks constructed by fMRI. \xl{One way to interpret important features is through looking at how the classification probability changes if the features are occluded or replaced. The major limitation of this approach is that replacing values may change the distribution of the data and lead to serious errors.} Therefore, we develop a 2-stage pipeline to eliminate the need to replace features for reliable biomarker interpretation. Specifically, we propose an inductive GNN to embed the graphs containing different properties of task-fMRI for identifying ASD and then discover the brain regions/sub-graphs used as evidence for the GNN classifier. We first \xl{show GNN can achieve high accuracy in identifying ASD}. Next, we calculate the feature importance scores using GNN and \xl{compare the interpretation ability with Random Forest}. Finally, we run with different atlases and parameters, proving the robustness of the proposed method. \xl{The detected biomarkers reveal their association with social behaviors and are consistent with those reported in the literature. We also show the potential of discovering new informative biomarkers}. Our pipeline can be generalized to other graph feature importance interpretation problems.
\end{abstract}

\section{Introduction}
Autism spectrum disorders (ASD) affect the structure and function of the brain. To better target the underlying roots of ASD for diagnosis and treatment, efforts to identify reliable biomarkers are growing \cite{goldani2014biomarkers}. Significant progress has been made using functional magnetic resonance imaging (fMRI) to characterize the brain remodeling in ASD \cite{Kaiser07122010}. % A common tool for representing fMRI data is graph, a kind of data structure which models a set of objects (nodes) and their relationships (edges). 
Recently, emerging research on Graph Neural Networks (GNNs) has combined deep learning with graph representation and applied an integrated approach to fMRI analysis in different neuro-disorders \cite{ktena2017distance}. Most existing approaches (based on Graph Convolutional Network (GCN) \cite{kipf2016semi}) require all nodes in the graph to be present during training and thus lack natural generalization on unseen nodes. Also, it is necessary to interpret the important feature in the data used as evidence for the model, but currently no tool exists that can interpret and explain GNNs while recent CNN explanation algorithms cannot directly work on graph input. %\One category of  explanation algorithms is \textit{Salience Mapping}, where a network repeatedly occludes portions of the input by replacing features to create a map showing how parts of the input affect the network output \cite{zeiler2014visualizing}.  Replacing the occluded features is challenging as the real data distribution is usually unknown. It may change the distribution of the original  data, causing serious errors, i.e. false positive errors.
 
\begin{figure}[t]
	\centering
	\centerline{\includegraphics[width=11cm]{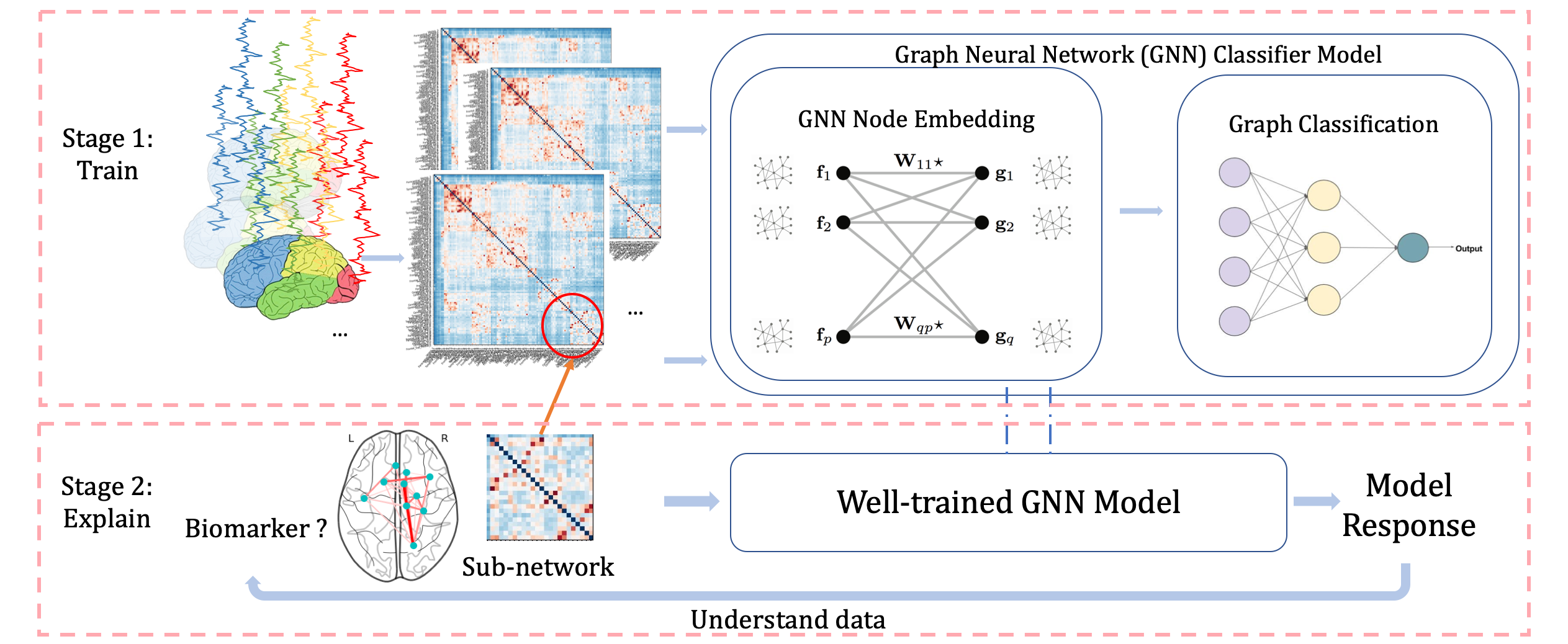}}
	%\centerline{(a) Result 1}\medskip
	\caption{Pipeline for interpreting important features from a GNN}
	\label{diagram}
\end{figure}

 Our main contributions include the following three points: 1) We develop a method to integrate all the available connectivity, geometric, anatomic information and task-fMRI (tfMRI) related parameters into graphs for deep learning. Our approach alleviates the problem of predetermining the best features and measures of functional connectivity, which is often ambiguous due to the intrinsic complex structure of task-fMRI. 2) We propose a generalizable GNN inductive learning model to more accurately classify ASD v.s.  healthy controls (HC).  Different from the spectral GCN \cite{kipf2016semi}, our GNN classifier is based on graph isomorphism, which can be applied to \xl{multigraphs} with different nodes/edges (e.g. sub-graphs)\xl{, and learn local graph information without binding to the whole graph structure}. 3) \xl{The GNN architecture enables us to train the model on the whole graph and validate it on subgraphs. We directly evaluate the importance scores on sub-graphs and nodes (i.e. regions of interest (ROIs)) by examining model responses, without resampling value for the occluded features.} The 2-stage pipeline to interpret important sub-graphs/ROIs, which are defined as biomarkers in our setting, is shown in Fig. \ref{diagram}.%This pipeline can be generalized to analyze different disease biomarkers as long as the data have an underlying graph structure.

\section{Methodology}
\subsection{Graph Definition}
\label{define}

We firstly parcellate the brain into $N$ ROIs based on its T1 structural MRI.  We define ROIs as graph nodes. We define an undirected multigraph $\bm{G} = (\bm{V},\bm{E})$, where $\bm{V} = ( \vec{v}_1,\vec{v}_2, \dots, \vec{v}_N)^T \in \mathbb{R}^{N\times D}$ and $\bm{E}= [\vec{e}_{ij}]\in \mathbb{R}^{N \times N \times F}$, \xl{$D$ and $F$ are the attribute dimensions of nodes and edges respectively}. For node attributes, we concatenate handcrafted features: degree of connectivity, General Linear Model (GLM) coefficients, mean, standard deviation of task-fMRI, and ROI center coordinates. We applied the Box-Cox transformation \cite{nishii2001box} to make each feature follow a normal distribution (parameters are learned from the training set and applied to the training and testing sets). The edge attribute $\bm{e}_{ij}$ of node $i$ and $j$ includes the Pearson correlation, partial correlation calculated using residual fMRI, and $exp(-r_{ij}/10)$ where $r_{ij}$ is the geometric distance between the centers of the two ROIs. \xl{We thresholded the edges under the 95th percentile of partial correlation values} to ensure sparsity for efficient computation and avoiding oversmoothing.

\subsection{Graph Neural Network (GNN) Classifier}

\begin{figure}[b]

	\centering
	\centerline{\includegraphics[width=12cm]{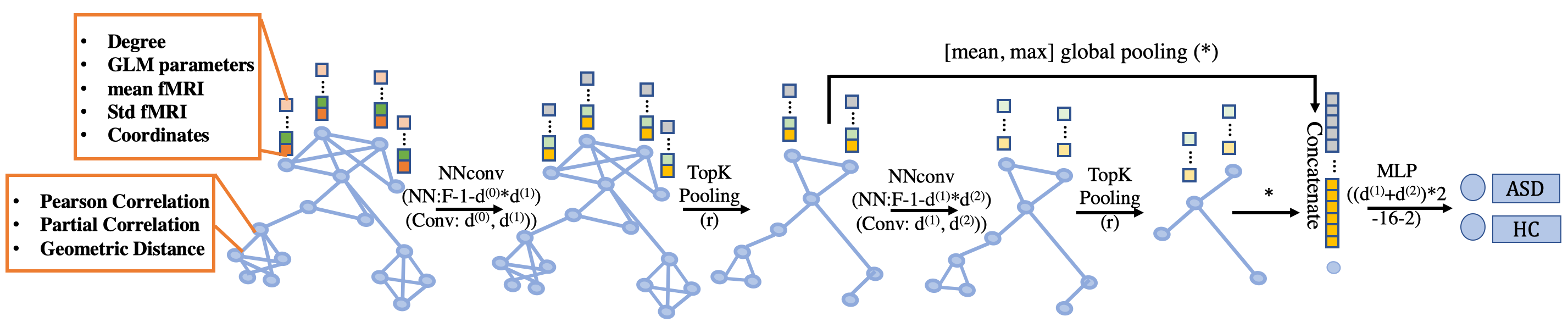}}
	\caption{The architecture of the GNN classifier}
	\label{network}

\end{figure}

The architecture of our proposed GNN is shown in Fig. \ref{network} (node, edge attribute definition, kernel sizes are denoted). The model inductively learns node representation by recursively aggregating and transforming feature vectors of its neighboring nodes. Below, we define the layers in the proposed GNN classifier.

\subsubsection{Convolutional Layer} \xl{Following} Message Passing Neural Networks (NNconv) \cite{gilmer2017neural}, which is invariant to graph symmetries, we leverage node degree in the embedding. The embedded representation of the $l$th convolutional layer $\vec {v}_{i}^{(l)} \in \mathbb{R}^{d^{(l)}}$ is 
\begin{equation}
    \vec {v}_{i}^{(l)}= \frac{1}{|\mathcal{N}(i)|+1}\sigma (\bm{\Theta} \vec{v}_{i}^{(l-1)} + \sum_{j \in \mathcal{N}(i)} h_{\phi}(\vec{e}_{ij}) \vec{v}_{j}^{(l-1)} ), 
\end{equation}
where $\sigma(\cdot)$ is a nonlinear activation function (we use \texttt{relu} here),  $\mathcal{N}(i)$ is node $i$'s 1-hop neighborhood, $\bm{\Theta} \in \mathbb{R}^{d^{(l)} \times d^{(l-1)}}$  is a learnable propagation matrix, $h_{\phi}$ denotes a Multi-layer Perceptron (MLP), which maps the edge attributes  $\bm{e}_{ij}$ to a $d^{(l)} \times d^{(l-1)}$ matrix, and we initialize $\vec{v}_{i}^{(0)} = \vec {v}_{i}$.

\subsubsection{Pooling Aggregation Layer}
To make sure that down-sampling layers behave idiomatically with respect to different graph sizes and structures, we adopt the approach in \cite{cangea2018towards} for reducing graph nodes. The choice of which nodes to drop is done based on projecting the node attributes on a learnable vector $\vec{w}^{{(l-1)}}\in\mathbb{R}^{d^{(l-1)}}$.
%In order to enable gradients to flow into $\vec{w}$, the projection scores are also used as gating values, suchthat retained 
The nodes receiving lower scores will experience less feature retention. Fully written out, the operation of this pooling layer (computing a pooled graph, $(\bm{V}^{(l)},\bm{E}^{(l)})$, from an input
graph, $(\bm{V}^{(l-1)}, \bm{E}^{(l-1)})$), is expressed as follows:
\begin{equation}
\resizebox{0.91\hsize}{!}{%
 $\vec{y} \!=\! \frac{\bm{V}^{(l-1)}{\vec{w}^{(l-1)}}}{\|\vec{w}^{(l-1)}\|} \quad \vec{i} \!=\! \text{top}k(\vec{y},k) \quad \bm{V}^{(l)}\! =\! (\bm{V}^{(l-1)} \odot \tanh(\vec{y}))_{\vec{i},:} \quad \bm{E}^{(l)} \!=\! \bm{E}^{(l-1)}_{\vec{i},\vec{i}}$%
 }.  
\end{equation}
Here $\|\cdot \|$ is the $L_2$ norm, top$k$ finds the indices corresponding to the largest $k$ elements in vector $\vec{y}$, $\odot$ is (broadcasted) element-wise multiplication, and $(\cdot)_{\vec{i},\vec{j}}$ is an indexing operation which takes elements at row indices specified by ${\vec{i}}$ and column indices specified by ${\vec{j}}$ (colon denotes all indices). The pooling operation trivially retains sparsity by requiring only a projection, a point-wise multiplication and a slicing into the original feature and adjacency matrix. Different from \cite{cangea2018towards}, we induce constraint  $\|\vec{w}^{(l)} \|_2 = 1$ implemented by adding an additional \texttt{regularization loss} $\lambda \sum_{l=1}^L (\|\vec{w}^{(l)} \|_2 - 1)^2$ to avoid identifiability issues.
\subsubsection{Readout Layer}
Lastly, we seek a “flattening” operation to preserve information about the input graph in a fixed-size representation.  Concretely, to summarise the output graph of the $l$th conv-pool block, $(\bm{V}^{(l)}, \bm{E}^{(l)})$, we use
\begin{equation}
    \vec{s}^{(l)} = (\frac{1}{N^{(l)}}\sum_{i=1}^{N^{(l)}}\vec{{v}}_i^{(l)})\parallel \max(\left\{\vec{{v}}_i^{(l)}:i=1,...,N^{(l)}\right\}),
\end{equation}
where $N^{(l)}$ is the number of graph nodes, $\vec{v}_i^{(l)}$ is the $i$th node's feature vector, $\max$ operates elementwisely, and $\parallel$ denotes concatenation. The final summary vector is obtained as the concatenation of all those summaries (i.e. $\vec{s}= \vec{s}^{(1)} \parallel\vec{s}^{(2)}\parallel \dots  \parallel \vec{s}^{(L)}$) and submitted to a MLP for obtaining final predictions.

\subsection{Explain Input Data Sensitivity}
To explain input data sensitivity, we cluster the whole brain graph into sub-graphs first. Then we investigate the predictive power of each sub-graph, further assign importance score to each ROI.

\subsubsection{Network Community Clustering}
\label{method1}
From now on we add the subscript to the graph as $\bm{G}_s = (\bm{V}_s,\bm{E}_s)$ for the $s$th instance, $s=1,...,S$, where $S$ is the number of graphs. Concatenating the sparsified non-negative partial correlation matrices $(\bm{E}_s)_{:,:,2}$ for all the graphs, we can create a 3rd-order tensor $\tau$ of dimension $N \times N \times S$.  Non-negative PARAFAC \cite{carroll1970analysis} tensor decomposition is applied to tensor $\tau$ to discover overlapping functional brain networks. Given decomposition rank $R$, $ \tau \approx \sum_{j = 1}^R \vec{a}_j\otimes\vec{b}_j\otimes\vec{c}_j$, where loading vectors $\vec{a}_j \in \mathbb{R}^{N}$, $\vec{b}_j \in \mathbb{R}^{N}$, $\vec{c}_j \in \mathbb{R}^{S}$ and $\otimes$ denotes the vector outer product. $\vec{a}_j = \vec{b}_j$ since the connectivity matrix is symmetric. The $i$th element of $\vec{a}_j$, $a_{ji}$ provides the membership of region $i$ in the community $j$. Here, we consider region $i$ belongs to community $j$ if $a_{ji} > mean(\vec{a}_j)+std(\vec{a}_j)$ \cite{loe2015comparison}. This gives us a collection of community indices indicating region membership $\left\{\vec{i}_j\subset\{1,...,N\}:j=1,...,R\right\}$.

\subsubsection{Graph Salience Mapping}
After decomposing all the brain networks into community sub-graphs $\left\{\bm{G}_{sj} = ((\bm{V}_s)_{\vec{i}_j,:},(\bm{E}_s)_{\vec{i}_j,\vec{i}_j}):
s=1,...,S,j=1,...,R\right\}$, we use a salience mapping method to assign each sub-graph an importance score. In our classification setting, the probability of class $c\in\{0,1\}$ (0: HC, 1: ASD) given the original network $\bm{G}$ is estimated from the predictive score of the model: $ p(c|\bm{G})$ . To calculate $p(c|\bm{G}_{sj})$, different from CNN or GCN, we can directly input the sub-graph into the pre-trained classifier. We denote $c_s$ as the class label for instance $s$ and define \textit{Evidence for Correct Class (ECC)} for each community: 
\begin{equation}
\label{score}
    \textit{ECC}_j = \frac{1}{S}\sum_{s} \tanh(\log_2(p(c=c_s|\bm{G}_{sj})/(1-p(c=c_s|\bm{G}_{sj})))),
\end{equation}
where laplace correction ($p \leftarrow (pS+1)/(S+2)$) is used to avoid zero denominators. \xl{Note that log odds-ratio is commonly used in logistic regression to make $p$ more separable}. The nonlinear tanh function is used for bounding \textit{ECC}. \textit{ECC} can be positive or negative. A positive value provides evidence for the classifier, whereas a negative value provides evidence against the classifier. The final importance score for node $k$ is calculated by $\sum_{j:k\in\vec{i}_j}\textit{ECC}_j/|\vec{i}_j|$. The larger the score, the more possible the node can be used as a distinguishable marker.

\section{Experiments and Results}
\subsection{Data Acquisition and Preprocessing}
We tested our method on a group of 75 ASD children and 43 age and IQ-matched healthy controls \xl{collected at Yale Child Study Center}. Each subject underwent a task fMRI scan (BOLD, TR = 2000 ms, TE = 25 ms, flip angle = $60^{\circ}$, voxel size $3.44\times3.44\times4\, mm^3$) acquired on a Siemens MAGNETOM Trio TIM 3T scanner.
For the fMRI scans, subjects performed the "biopoint" task, viewing point light animations of coherent and scrambled biological motion in a block design \cite{Kaiser07122010} ($24s$ per block). The fMRI data was preprocessed following the pipeline in \cite{yang2016brain}.  

\xl{The mean time series for each node were extracted from a random $1/3$ of voxels in the ROI (given an atlas) of preprocessed images by bootstrapping}. We augmented the ASD data 10 times and the HC data 20 times, resulting in 750 ASD graphs and 860 HC graphs separately. We split the data into 5 folds based on subjects. Four folds were used as training data and the left out fold was used for testing. Based on the definition in Section \ref{define}, each node attribute $\vec{v}_i \in \mathbb{R}^{10}$ and each edge attribute $\vec{e}_{ij} \in \mathbb{R}^3$.  Specifically, the GLM parameters of "biopoint task" are: $\beta_1:$ coefficient of biological motion matrix; $\beta_3$: coefficient of scramble motion matrix; $\beta_2$ and $\beta_4$:  coefficients of  the previous two matrices' derivatives.

\begin{table}[t]	
	\centering
	\caption{Performance of different models (mean$\pm$ std)}
	%(\emph{Input =2-channel}, $w=3$, $stride =1$)
	\scalebox{0.8}{
	\begin{tabular}{p{2cm}|c|c|c|c|c|c}
		\hline
		\textbf{Model}&RF(V)&RF(E)&RF(V+E)&GNN(r=0.3)& GNN(r=0.5)&GNN(r=0.8)\\ \hline
		\textbf{Accuracy}&$0.71\!\pm\!0.05$&$0.66\!\pm\!0.06$&$0.68\!\pm\!0.06$&$0.67\!\pm\!0.14$&$\bm{0.76\!\pm\!0.06}$&$0.73\!\pm\!0.07$\\
		\textbf{F-score}&$0.69\!\pm\!0.06$&$0.68\!\pm\!0.06$&$0.63\!\pm\!0.12$&$0.68\!\pm\!0.09$&$\bm{0.79\!\pm\!0.08}$&$0.71\!\pm\!0.10$\\
		\textbf{Precision}&$0.68\!\pm\!0.06$&$0.61\!\pm\!0.06$&$0.69\!\pm\!0.12$&$0.65\!\pm\!0.19$&$\bm{0.76\!\pm\!0.12}$&$0.68\!\pm\!0.08$\\
		\textbf{Recall}&$0.73\!\pm\!0.12$&$0.76\!\pm\!0.10$&$0.77\!\pm\!0.09$&$0.74\!\pm\!0.07$&$\bm{0.82\!\pm\!0.06}$&$0.75\!\pm\!0.08$\\
		\hline 
	\end{tabular}
	\label{tb3}	
}
\end{table}
\subsection{Step 1: Train ASD/HC Classification Model}
\label{classification}

 Firstly, we tested classifier performance on the Destrieux atlas \cite{destrieux2010automatic} (148 ROIs) using proposed GNN. \xl{Since our pipeline integrated interpretation and classification, we apply a random forest (RF) using 1000 trees  as an additional "reality check", while the other existing graph classification models either cannot achieve the performance as GNN \cite{cangea2018towards,gilmer2017neural} or do not have straightforward and reliable interpretation ability \cite{adebayo2018sanity}.} We flattened the features to $\bm{V}\in \mathbb{R}^{1480}$ and $\bm{E}\in \mathbb{R}^{65712}$ ($65712 = 148 \times 148 \times 3$) and used this input to the RF.  In our GNN, \xl{$d^{(0)}=D=10$, $d^{(1)} = 16, d^{(2)} = 8$, resulting in 2746 trainable parameters} and we tried different pooling ratios $r$ ($k = r \times N)$ in Fig. \ref{network}, which was implemented based on \cite{Fey/Lenssen/2019}.  We applied \texttt{softmax} after the network output and combined \texttt{cross entropy loss} and \texttt{regularization loss} with $\lambda = 0.001$ as the objective function. We used the Adam optimizer with initial learning 0.001, then decreased it by a factor of $10$ every 50 epochs. We trained the network 300 epochs for all of the splits and measured the instance classification by accuracy, F-score, precision and recall (see Table \ref{tb3}). Our proposed model significantly outperformed the alternative method, due to its ability to embed high dimensional features based on the structural relationship. We selected the best GNN model with  $r=0.5$  in the next step: interpreting biomarkers.

\subsection{Step 2: Interpret and Explain Biomarkers}
\begin{figure}[b]
	\centering
	\centerline{\includegraphics[width=12cm]{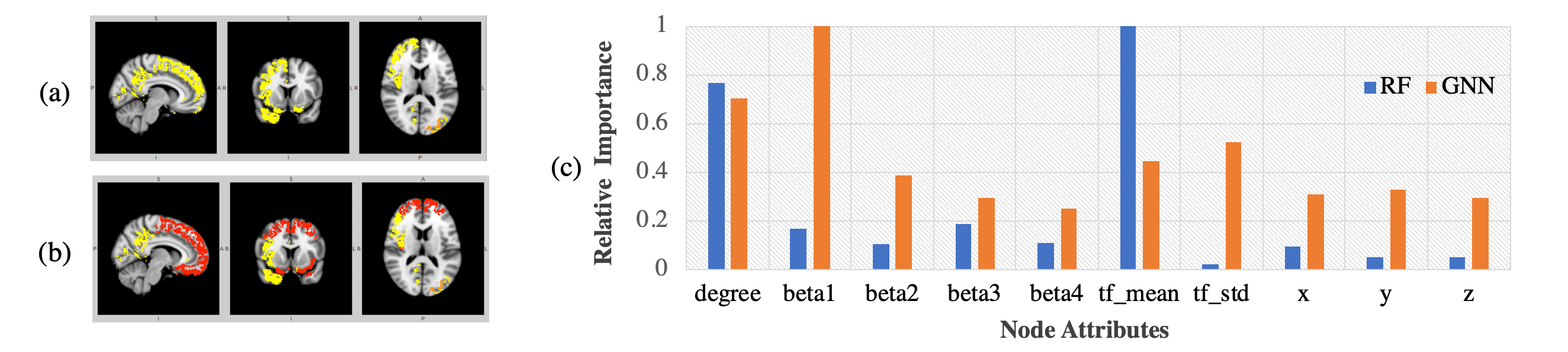}}
	%\centerline{(a) Result 1}\medskip
	\caption{(a) Top 30 important ROIs (colored in yellow) selected by RF; (b) Top 30 important ROIs selected by GNN ($R$=20) (colored in red) laying over (a); (c) Node attributes' relative importance scores in the two methods. }
	\label{RF}
\end{figure}
\begin{figure}[t]
	\centering
	\centerline{\includegraphics[width=12cm]{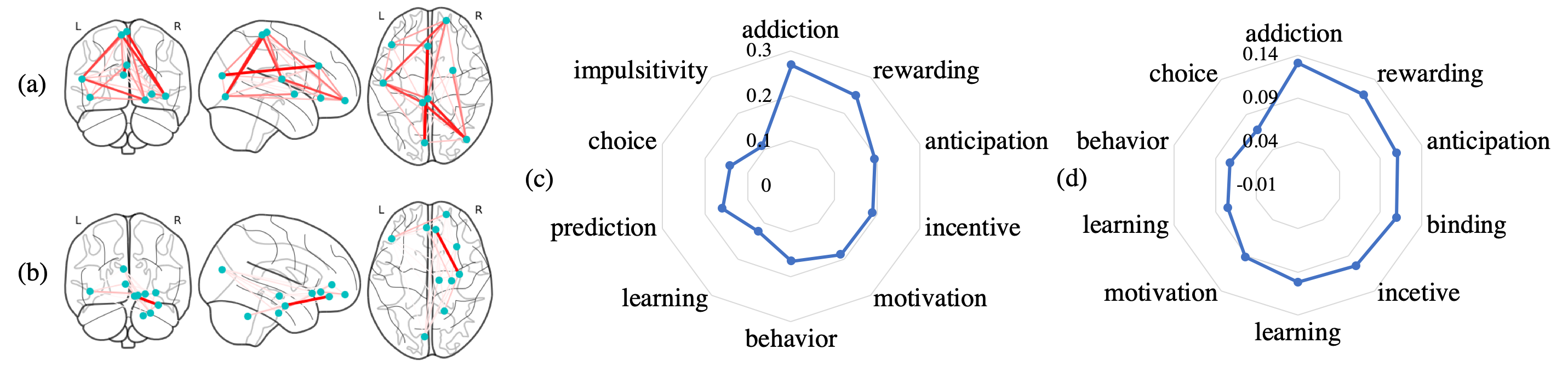}}
	%\centerline{(a) Result 1}\medskip
	\caption{(a) (c) Top scoring sub-graph and corresponding functional decoding keywords and coefficients. (b) (d) The 2nd high scoring sub-graph and corresponding functional decoding keywords and coefficients.  }
	\label{res1-2}
\end{figure}

We put forth the hypothesis that the more accurate the classifier, the more reliable biomarkers can be found. We used the best RF model using $\bm{V}$ as inputs ($77.4\%$ accuracy on testing set) and used the RF-based feature importance (mean Gini impurity decrease) as a form of standard method for comparison.  For GNN interpretation, we also chose the best model ($83.6\%$ accuracy on testing set). Further, to be comparable with RF, all of the interpretation experiments were performed on the training set only. The interpretation results are shown in Fig.~\ref{RF}, where the top 30 important ROIs (averaged over node features and instances) selected by RF are shown in yellow and the top 30 important ROIs selected by our proposed GNN in red. Nine important ROIs were selected by both methods. In addition, for node attribute importance, we averaged the importance score over ROIs and instances for RF. For GNN, we averaged \textit{gradient explanation} over all the nodes and instances, i.e. $\mathbb{E}(\frac{1}{N}\sum_i|\frac{\partial y}{\partial v_{ij}}|)$, where $y = p(c=1|\bm{G})$, which quantifies \xl{the sensitivity of the $j$th node attribute}. From Fig. \ref{RF}(c) we show the relative importance to the most important node attribute, our proposed method assigned more uniform importance to each node attribute, among which the biological motion parameter $\beta_1$ was the most important. %And degree's importance was less significant, since GNN can summarize it from graph edges.  
In addition, similar features, mean/std of task-fMRI (tf\_mean/tf\_std) and coordinates $(x,y,z)$, have similar scores, which makes more sense for human interpretation.  %among which degree's importance is not so significant, since GNN can summarize it from graph edges. In addition, similar features: GLM parameters and coordinates $(x,y,z)$ have almost equal scores, which makes more sense for human interpretation. 
Notice that our proposed pipeline is also able to identify sub-graph importance from Eq.~(\ref{score}), which is helpful for understanding the interaction between different brain regions. We selected the top 2 sub-graphs ($R$=20) and used Neurosynth  \cite{yarkoni2011large} to decode the functional keywords associated with the sub-graphs (shown in Fig. \ref{res1-2}). These networks are both associated with high-level social behaviors. To illustrate the predictive power of the 2 sub-graphs, we retrained the network using the graph slicing on those 19 ROIs of the 2 sub-graphs as input. Accuracy on the testing set (in the split of the best model) was $78.9\%$, achieving comparable performance to using the whole graph.

\subsection{Evaluation: Robustness Discussion}
\begin{figure}[t]
	\centering
	\centerline{\includegraphics[width=12cm]{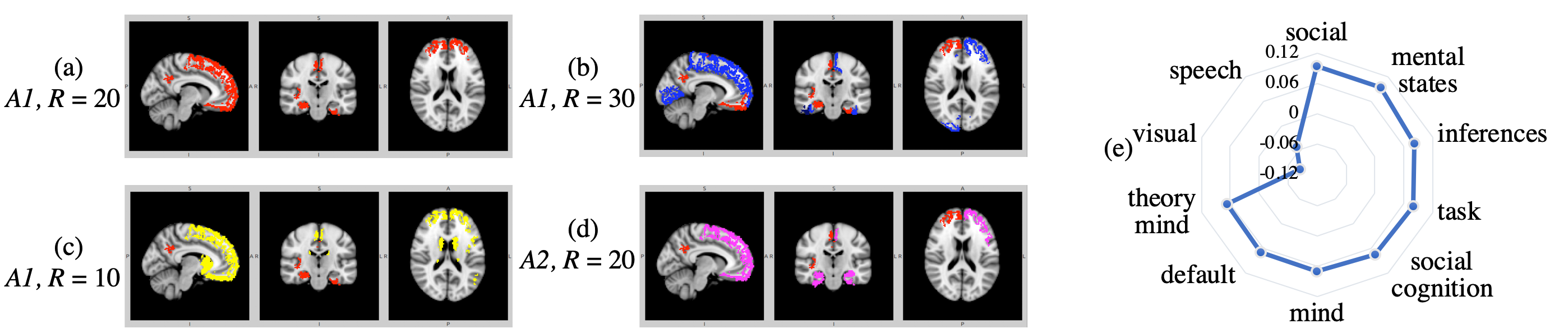}}
	%\centerline{(a) Result 1}\medskip
	\caption{(a) The biomarkers (red) interpreted on \textit{A1} with 20 clusters; (b)-(d) The biomarkers interpreted by different $R$ and altas laying over on (a) with different colors; (e) The correlation between overlapped ROIs and functional keywords.}
	\label{rob}
\end{figure}

To examine the potential influence of different graph building strategies on the reliability of network estimates, the functional and anatomical data were registered and parcellated by the \xl{Destrieux atlas  (\textit{A1}) and the Desikan-Killiany atlas (\textit{A2}) \cite{desikan2006automated}}.  We also showed the robustness of the results with respect to the number of clusters for $R = 10, 20, 30$. The results are shown in Fig. \ref{rob}. We ranked \textit{ECC}s for each node and indicated the top 30 ROIs in \textit{A1} and top 15 ROIs in \textit{A2}. The altas and number of clusters are indicated on the left of each sub-figure. Orbitofrontal cortex and ventromedial prefrontal cortex are selected in all the cases, which are social motivation related and have previously been shown to be associated with ASD \cite{Kaiser07122010}.  We also validated the results by decoding the neurological functions of the important ROIs overlapped with Neurosynth. %More interpretation results are  shown in supplementary.

\section{Conclusion and Future Work}
In this paper, we proposed a  framework to discover ASD brain biomarkers from task-fMRI using GNN. \xl{It achieved improved accuracy and more interpretable features than the baseline method.} We also showed our method performed robustly on different atlases and hyper-parameters. Future work will
include investigating more hyper-parameters (i.e. suitable size of sub-graphs
communities), testing the results on functional atlases and different graph definition methods. The pipeline can be generalized to other feature importance analysis problems, such as resting-fMRI biomarker discovery and vessel cancer detection.

\section*{Acknowledgments}
This work was supported by NIH Grant R01 NS035193.

\bibliographystyle{splncs04}
\bibliography{ref}

\end{document}